# A New Look at the Classical Entropy of Written English

Submitted to the IEEE Transactions on Information Theory

Fabio G. Guerrero

Electrical and Electronics Engineering School, Universidad del Valle, Cali, Colombia (South America).

Calle 13 100-00, Ciudad Universitaria Meléndez, Edificio 355, Oficina 2005, Cali, Colombia.

Phone: (+572) 3392140 ext 109

e-mail: fguerrer@univalle.edu.co

*Abstract*

A simple method for finding the entropy and redundancy of a reasonable long sample of English text by direct computer processing and from first principles according to Shannon theory is presented. As an example, results on the entropy of the English language have been obtained based on a total of 20.3 million characters of written English, considering symbols from one to five hundred characters in length. Besides a more realistic value of the entropy of English, a new perspective on some classic entropy-related concepts is presented. This method can also be extended to other Latin languages. Some implications for practical applications such as plagiarism-detection software, and the minimum number of words that should be used in social Internet network messaging, are discussed.

## I. INTRODUCTION

Finding a precise value of the entropy of a language is, in general, an elusive matter, mainly due to the underlying statistical nature of the problem. The entropy of English has being researched for more than sixty years by a variety of approaches. Yet coincidence is not perfect among reported values of entropy. There may be different assumptions, of course, but most of the previous work has been based on indirect methods.

To have a reliable estimate of the entropy of a language is something important. For instance, coding methods in source coding theory aim to code as close as possible to the entropy limit. If a good estimation of entropy is available, it may help to know how good a system of source coding is.

Information Theory's father, C. E. Shannon showed in [1] that the entropy of printed English should be bounded between 0.6 and 1.3 bits/letter over 100-letter long sequences of English text. He used for his calculations a human prediction approach and ignored punctuation, lowercase, and uppercase. Cover and King, using gambling estimates, found a value between 1.25 and 1.35 bits per character [2]. Teahan and Cleary reported a value for the entropy of English of 1.46 bits per character using data compression [3]. Kontoyiannis reported a value of 1.77 bits per character using a method based on string matching, resembling that of universal source coding, based on a one-million-character sample from a same author [4]. Stylistics language analysis based on both frequency of words and entropic analysis is also a wide field of research. A recent example of this kind of work is presented in [5].

In this paper, an approach for estimating the entropy rate from several English samples using direct computer calculations of probability and from entropy first principles is presented. We use C. E. Shannon's mathematical analysis for calculating the entropy $H$ of a natural language by a series of approximations of conditional entropy $F_0$, $F_1$, $F_2$,… given by

$$F_N = -\sum_{i,j} p(B_i, j) \log_2 p(j, B_i) + \sum_i p(B_i) \log_2 p(B_i) \qquad (1)$$

In (1), $B_i$ is a block of $N$-1 characters, $j$ is the next character after $B_i$, $p(B_i, j)$ is the probability of the $N$-character block $(B_i, j)$, and $p(j \mid B_i)$ is the conditional probability of character $j$ after block $B_i$. The language entropy $H$, in bits/character, can be calculated as:

$$H = \operatorname*{Lim}_{N \to \infty} F_N \qquad\qquad (2)$$

Section 2 of this paper discusses the methodology (software, practical methods, assumptions, etc.) adopted to obtain the entropy values reported in this work. Section 3 presents the main results obtained. In section 4, a discussion of the results is given and, finally, section 5 summarizes the main findings of this work. This paper concentrates exclusively on the entropy of English at a probabilistic level. Analysis of the language at any other level is beyond the scope of this paper. All the evidence material supporting this work is available in [6].

## II. METHODOLOGY

Twenty-one classic literary works are used as the basis for calculations in this work. These samples were selected among the top novels reported by The Daily Telegraph British newspaper on January 2009 in an article entitled "100 novels everyone should read" [7].

The text of the selected works, all of them in contemporary English, is freely available on the Internet at the Guternberg project homepage [8]. For those few works whose text for some reason was not available, a different work but from the same author was chosen. The resulting selection is quite diverse in terms of author's origin, literary style, and such like.

All the software used in this work was written in Mathematica™ 6. Computer processing was carried out on a laptop computer with an AMD 1.6 GHz double-core processor and 1024 MB physical memory. Table I shows the twenty-one samples selected and their basic statistics.

TABLE I
SELECTED SAMPLES OF ENGLISH LANGUAGE

| Sample | Title | Author | Number of characters | Alphabet size | Number of words | Different words | WDR(%) | $\alpha$ |
|---|---|---|---|---|---|---|---|---|
| 1 | Don Quixote | Miguel de Cervantes | 2158307 | 71 | 414088 | 16357 | 3.95 | 4.14 |
| 2 | Anna Karenina | Leo Tolstoy | 1933480 | 87 | 358681 | 13888 | 3.87 | 4.25 |
| 3 | David Copperfield | Charles Dickens | 1905581 | 76 | 363885 | 15477 | 4.25 | 4.07 |
| 4 | Middlemarch | George Eliot | 1760448 | 79 | 323641 | 16403 | 5.07 | 4.32 |
| 5 | Ulysses | James Joyce | 1488035 | 96 | 272357 | 34287 | 12.59 | 4.36 |
| 6 | Moby Dick | Herman Melville | 1196272 | 78 | 218522 | 18976 | 8.68 | 4.36 |
| 7 | The Moonstone | Wilkie Collins | 1053029 | 79 | 199982 | 9791 | 4.90 | 4.16 |
| 8 | Tristram Shandy | Laurence Sterne | 1010435 | 77 | 190890 | 14795 | 7.75 | 4.16 |
| 9 | Jane Eyre | Charlotte Brontë | 1009233 | 79 | 189448 | 13555 | 7.15 | 4.16 |
| 10 | Night and Day | Virginia Woolf | 928473 | 73 | 170554 | 10952 | 6.42 | 4.34 |
| 11 | Tess of the Durbervilles | Thomas Hardy | 820274 | 74 | 153128 | 12742 | 8.32 | 4.24 |
| 12 | Pride and Prejudice | Jane Austen | 671721 | 74 | 122868 | 6832 | 5.56 | 4.37 |
| 13 | The Portrait of a Lady (Part I) | Henry James | 643863 | 70 | 121693 | 8680 | 7.13 | 4.19 |
| 14 | Madame Bovary | Gustave Flaubert | 638533 | 75 | 117654 | 10930 | 9.29 | 4.32 |
| 15 | Wuthering Heights | Emily Brontë | 637966 | 72 | 119403 | 9803 | 8.21 | 4.17 |
| 16 | The Portrait of a Lady (Part II) | Henry James | 615365 | 69 | 118158 | 7484 | 6.33 | 4.10 |
| 17 | Robinson Crusoe | Daniel Defoe | 522033 | 78 | 100922 | 6144 | 6.09 | 4.09 |
| 18 | Right Ho Jeeves | PG Wodehouse | 394233 | 80 | 76148 | 8009 | 10.52 | 4.03 |
| 19 | Cranford | Elizabeth Gaskell | 379153 | 76 | 72396 | 6840 | 9.45 | 4.14 |
| 20 | The War of the Worlds | HG Wells | 332769 | 75 | 60897 | 7239 | 11.89 | 4.37 |
| 21 | Heart of Darkness | Joseph Conrad | 207403 | 69 | 39094 | 5743 | 14.69 | 4.18 |

In Table I, $\alpha$ is the average word length, $\sum L_i p_i$ , where $L_i$ is the length in characters of the $i$-th word, and $p_i$ is its corresponding probability. In Table I the weighted average value of $\alpha$ is 4.22 letters per word, the total number of words is 3,804,409 and the total number of characters is 20,306,606. The number of characters corresponds only to printable characters (i.e. control characters not included). The word dispersion ratio, WDR, equals the number of different words over the total number of words.

Because, with the exception of control characters, all printable characters present in each sample text are taken into account, uppercase and lowercase characters are different, punctuation is included, the space character is included, etc.

Entropy for symbols (blocks) of $n = 1$ to 500 characters in length is calculated using the fundamental formula $H = -\sum_{i} p_i \log_2 p_i$. The probability $p_i$ of the $n$-character symbol $B_i$ *is* obtained using the law of large numbers, where $p(B_i) \approx$ *number of occurrences of $B_i$ /total number of symbols.*

To get the entropy for symbols longer than one character in length, entropy has to be averaged over individual shifted symbol entropies for the same $n$. That is, the $n$-character symbol probability is computed by first counting non-overlapping symbols from the first character of the text (*shift* = 0). Then, the $n$-character symbol probability is computed for non-overlapping symbols starting from the second character (*shift* = 1), and so on. It can be easily observed that the number of shifts that have to be considered before $n$-character blocks start repeating is equal to $n$. Remarkably, the shift entropies of a given $n$ are very similar. For example, with sample 11 (*Tess of the Durbervilles*), for $n = 5$ (blocks of five characters), the five entropy values (*shifts* from 0 to 4) obtained are 14.256508, 14.257813, 14.248730, 14.256469, and 14.251314 bits/symbol. This same behavior is observed when the analysis is done using words instead of characters. The same approach using $n$-character symbols, however, leads to a finer analysis because it avoids the quantization in length imposed by words, and also considers non-alphanumeric characters, which are part of the language.

Also, for every sample, a variable named equiprobability distance, $n_{aep}$, is calculated. The value of $n_{aep}$ is such that, for any $n \geq n_{aep}$, equiprobable symbols are obtained for all shifts of $n$.

## III. Results

### A. n-character symbol entropy

For reasons of space, Table II shows the values of entropy for $n$-character symbols for $n = 1$ to 15 characters only. The values in Table II have been rounded to two significant digits. The complete table ($n = 1$ to 500) is available at [6]. In Table II, an asterisk indicates the maximum $n$-character symbol entropy of every sample.

TABLE II
E NTROPY VALUES FOR $n$-CHARACTER SYMBOLS

| Sample | Block Length ($n$) | | | | | | | | | | | | | | |
|---|---|---|---|---|---|---|---|---|---|---|---|---|---|---|---|
| | 1 | 2 | 3 | 4 | 5 | 6 | 7 | 8 | 9 | 10 | 11 | 12 | 13 | 14 | 15 |
| 1 | 4.39 | 7.85 | 10.59 | 12.72 | 14.34 | 15.53 | 16.36 | 16.88 | 17.17 | 17.29 | 17.32* | 17.29 | 17.24 | 17.17 | 17.09 |
| 2 | 4.41 | 7.94 | 10.67 | 12.75 | 14.31 | 15.45 | 16.23 | 16.73 | 17.00 | 17.13 | 17.15* | 17.13 | 17.07 | 17.00 | 16.92 |
| 3 | 4.43 | 7.97 | 10.74 | 12.85 | 14.44 | 15.60 | 16.37 | 16.84 | 17.09 | 17.18 | 17.19* | 17.15 | 17.08 | 17.01 | 16.92 |
| 4 | 4.40 | 7.94 | 10.72 | 12.83 | 14.41 | 15.53 | 16.29 | 16.74 | 16.98 | 17.08 | 17.09* | 17.04 | 16.98 | 16.90 | 16.81 |
| 5 | 4.55 | 8.22 | 11.22 | 13.58 | 15.22 | 16.23 | 16.78 | 17.03 | 17.09* | 17.06 | 16.98 | 16.89 | 16.78 | 16.69 | 16.59 |
| 6 | 4.43 | 7.99 | 10.87 | 13.06 | 14.62 | 15.65 | 16.26 | 16.57 | 16.69 | 16.69* | 16.64 | 16.56 | 16.46 | 16.37 | 16.27 |
| 7 | 4.45 | 7.95 | 10.69 | 12.72 | 14.18 | 15.18 | 15.81 | 16.17 | 16.33 | 16.38* | 16.37 | 16.31 | 16.24 | 16.16 | 16.07 |
| 8 | 4.44 | 7.96 | 10.78 | 12.90 | 14.38 | 15.34 | 15.92 | 16.22 | 16.35 | 16.36* | 16.32 | 16.26 | 16.18 | 16.09 | 16.01 |
| 9 | 4.43 | 7.98 | 10.81 | 12.97 | 14.52 | 15.53 | 16.12 | 16.40 | 16.49* | 16.48 | 16.41 | 16.32 | 16.22 | 16.13 | 16.03 |
| 10 | 4.39 | 7.88 | 10.62 | 12.67 | 14.12 | 15.10 | 15.71 | 16.05 | 16.21 | 16.25* | 16.22 | 16.16 | 16.08 | 15.99 | 15.90 |
| 11 | 4.39 | 7.90 | 10.68 | 12.78 | 14.25 | 15.22 | 15.79 | 16.07 | 16.17* | 16.16 | 16.10 | 16.02 | 15.92 | 15.83 | 15.73 |
| 12 | 4.41 | 7.89 | 10.57 | 12.51 | 13.87 | 14.78 | 15.34 | 15.64 | 15.77 | 15.80* | 15.76 | 15.70 | 15.61 | 15.53 | 15.44 |
| 13 | 4.43 | 7.93 | 10.65 | 12.64 | 14.02 | 14.90 | 15.43 | 15.68 | 15.77 | 15.77* | 15.72 | 15.64 | 15.56 | 15.47 | 15.38 |
| 14 | 4.39 | 7.88 | 10.67 | 12.74 | 14.14 | 15.01 | 15.51 | 15.75 | 15.82* | 15.81 | 15.74 | 15.65 | 15.56 | 15.46 | 15.37 |
| 15 | 4.42 | 7.95 | 10.72 | 12.79 | 14.21 | 15.10 | 15.59 | 15.81 | 15.86* | 15.82 | 15.75 | 15.66 | 15.56 | 15.46 | 15.37 |
| 16 | 4.42 | 7.90 | 10.57 | 12.51 | 13.87 | 14.76 | 15.29 | 15.57 | 15.68 | 15.69* | 15.65 | 15.58 | 15.49 | 15.40 | 15.31 |
| 17 | 4.29 | 7.67 | 10.28 | 12.20 | 13.55 | 14.45 | 15.01 | 15.31 | 15.43 | 15.45* | 15.41 | 15.34 | 15.26 | 15.17 | 15.07 |
| 18 | 4.49 | 7.99 | 10.72 | 12.66 | 13.92 | 14.65 | 15.02 | 15.17 | 15.19* | 15.14 | 15.05 | 14.96 | 14.86 | 14.76 | 14.67 |
| 19 | 4.39 | 7.91 | 10.63 | 12.55 | 13.79 | 14.53 | 14.92 | 15.09 | 15.11* | 15.07 | 14.99 | 14.90 | 14.81 | 14.71 | 14.62 |
| 20 | 4.36 | 7.85 | 10.56 | 12.49 | 13.71 | 14.41 | 14.78 | 14.93 | 14.95* | 14.90 | 14.82 | 14.72 | 14.62 | 14.53 | 14.43 |
| 21 | 4.40 | 7.90 | 10.63 | 12.51 | 13.61 | 14.16 | 14.39 | 14.43* | 14.38 | 14.29 | 14.18 | 14.06 | 13.95 | 13.85 | 13.75 |

The values of entropy, in Table II, are averages of the individual shifted entropies. The total number of individual shifted entropies computed for every sample, from $n = 1$ to $k$, is $\frac{k(k+1)}{2}$. Then, for $n = 1$ to 500, this number is 125,250. For ease of display, Fig. 1 shows only the curves of $H_n$ for four samples, from $n = 1$ to 100.

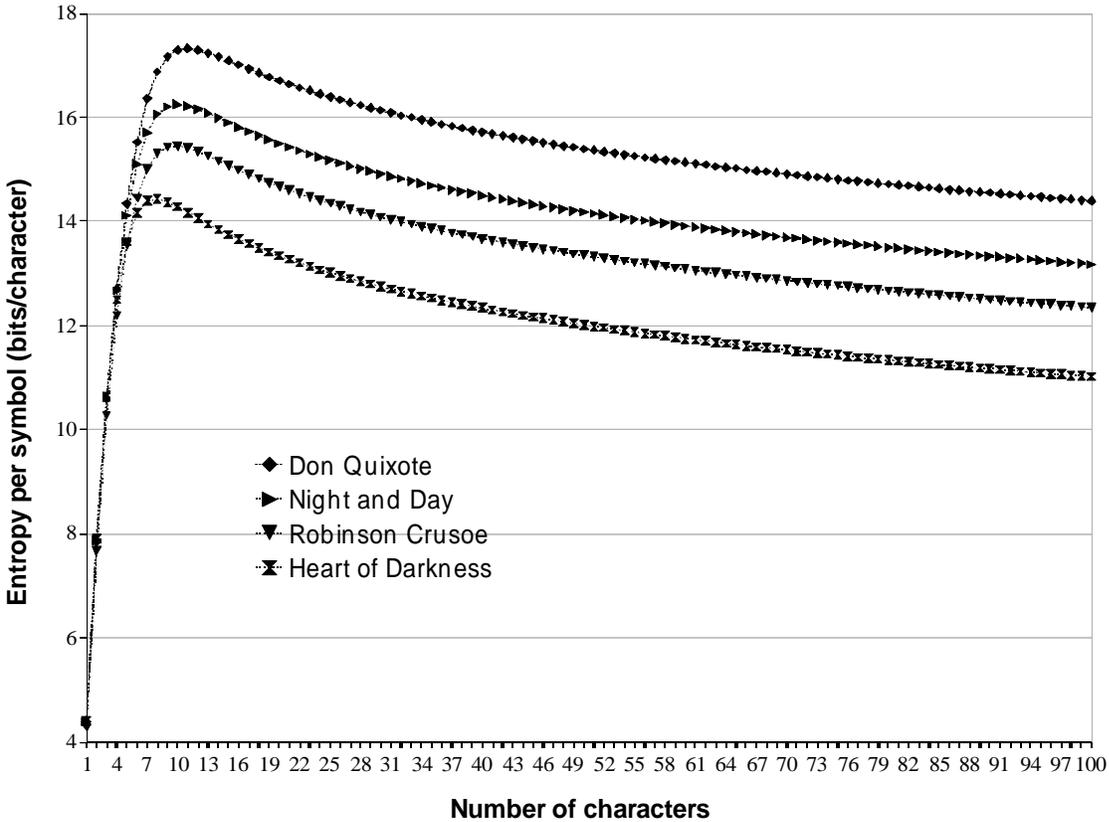

Fig. 1. Entropy values for *n* = 1 to 100 character symbols.

*B. CPU time used*

Fig. 2 shows the approximate total CPU time used on every sample to get the values reported, as in Table I, for *n* = 1 to 500. The algorithm employed for obtaining the frequency of symbols is an extremely simple sorting algorithm implemented with Mathematica functions. In Fig. 2, the CPU time is the sum of the individual CPU times (as reported by Mathematica's function *Timing*) for every entropy calculation. The work is carried out simultaneously on both cores of the computer's processor. Half of the samples are processed on core 1, and the other half on core 2, balancing the load, as given by every sample's number of characters, as much as possible. Hence, the actual real time needed to process the whole set of samples is nearly half the total sum of individual CPU times. An analysis based on computational complexity theory is beyond the scope of this paper.

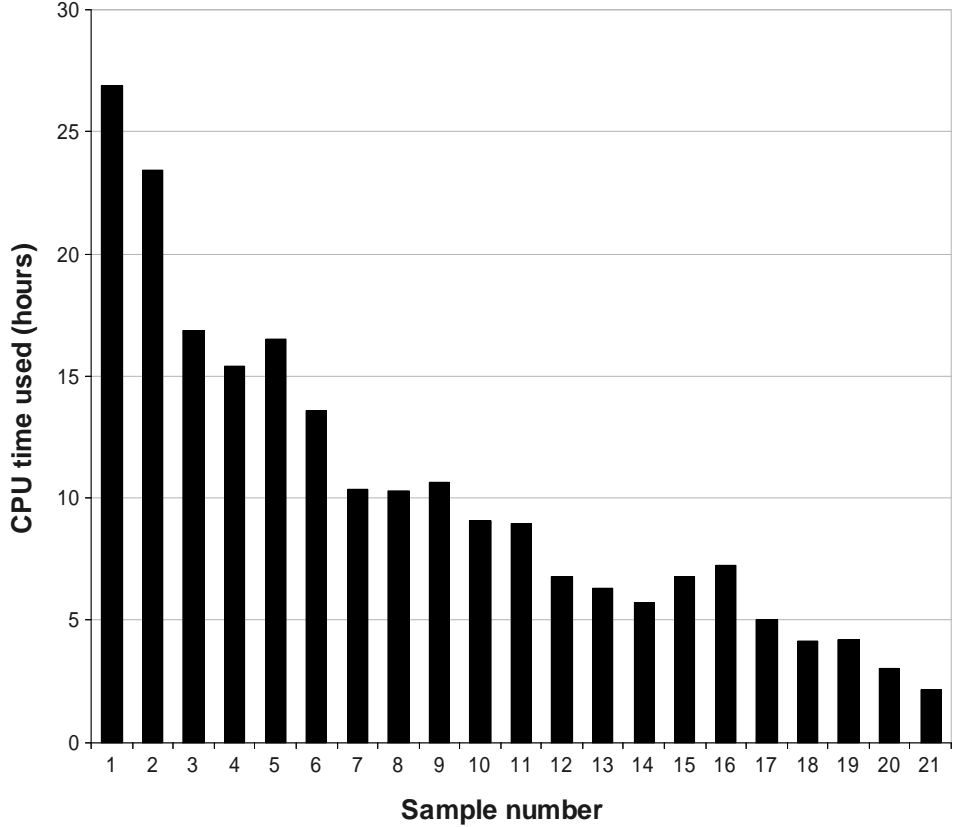

Fig. 2. Approximate CPU time used.

*C. Equiprobability distance*

Table III shows the values of $n_{aep}$ for the twenty-one samples, for $n$ between 1 and 200. As mentioned, $n_{aep}$ is the length in characters for which for any $n$ greater or equal than $n_{aep}$ the $n$-character symbols are equiprobable, that is $H = log_2 \left\lfloor \frac{Total\ number\ of\ characters - shift}{n} \right\rfloor$, for *all* shifts (0 to $n$-1) within a given $n$. This definition of $n_{aep}$ is quite stringent since it requires equiprobability for every shift and for every $n \geq n_{aep}$. For example, on sample 11 there exist symbol equiprobability from $n = 76$ to 307 for every shift, but clearly 75 does not equal $n_{aep}$. If there were repeated subsequences of text greater than 500 characters on any sample, something which is very unlikely with a reasonable degree of probability, then the values in Table III would be absolutely certain only for $n \leq 500$. Some implications of $n_{aep}$ will be discussed in the next section.

TABLE III
SYMBOL LENGTH FOR EQUIPROBABLE
SYMBOLS ($1 \leq n \leq 500$)

| Sample | $n_{aep}$ |
|--------|-----------|
| 1 | 130 |
| 2 | 49 |
| 3 | 73 |
| 4 | 47 |
| 5 | 99 |
| 6 | 83 |
| 7 | 61 |
| 8 | 69 |
| 9 | 57 |
| 10 | 38 |
| 11 | 308 |
| 12 | 41 |
| 13 | 43 |
| 14 | 35 |
| 15 | 27 |
| 16 | 36 |
| 17 | 34 |
| 18 | 46 |
| 19 | 30 |
| 20 | 36 |
| 21 | 75 |

## IV. DISCUSSION

### A. Conditional Entropy

$F_0$ equals $\log_2(A_S)$, where $A_S$ is the alphabet size (fifth column on Table I). $F_1$ is calculated from single-character frequencies as given by $F_1 = -\sum_i p(i) \log_2 p(i)$ bits per character. Following (2), $F_2$ is given by

$$F_2 = -\sum_{i,j} p(i,j) \log_2 p(i,j) + \sum_i p(i) \log_2 p(i) \qquad (3)$$

Similarly, $F_3$ is given by

$$F_3 = -\sum_{i,j,k} p(i,j,k) \log_2 p(i,j,k) + \sum_{i,j} p(i,j) \log_2 p(i,j) \qquad (4)$$

and so on. Table II shows values for $F_N$ from $F_1$ to $F_{15}$, rounded to three significant digits.

TABLE IV
ENTROPY VALUES FOR $F_N$

| Sample | $N$ | | | | | | | | | | | | | | |
|---|---|---|---|---|---|---|---|---|---|---|---|---|---|---|---|
| | 1 | 2 | 3 | 4 | 5 | 6 | 7 | 8 | 9 | 10 | 11 | 12 | 13 | 14 | 15 |
| 1 | 4.392 | 3.455 | 2.747 | 2.123 | 1.622 | 1.194 | 0.826 | 0.522 | 0.285 | 0.128 | 0.028 | -0.028 | -0.056 | -0.072 | -0.077 |
| 2 | 4.414 | 3.521 | 2.734 | 2.078 | 1.563 | 1.139 | 0.785 | 0.494 | 0.273 | 0.126 | 0.028 | -0.026 | -0.058 | -0.071 | -0.077 |
| 3 | 4.426 | 3.543 | 2.773 | 2.111 | 1.592 | 1.152 | 0.775 | 0.468 | 0.246 | 0.099 | 0.009 | -0.042 | -0.067 | -0.078 | -0.083 |
| 4 | 4.398 | 3.538 | 2.783 | 2.111 | 1.577 | 1.127 | 0.754 | 0.455 | 0.237 | 0.097 | 0.009 | -0.041 | -0.067 | -0.080 | -0.083 |
| 5 | 4.551 | 3.666 | 3.007 | 2.351 | 1.644 | 1.009 | 0.548 | 0.251 | 0.066 | -0.031 | -0.077 | -0.097 | -0.102 | -0.100 | -0.095 |
| 6 | 4.427 | 3.565 | 2.874 | 2.190 | 1.565 | 1.027 | 0.611 | 0.311 | 0.116 | 0.008 | -0.055 | -0.082 | -0.094 | -0.095 | -0.094 |
| 7 | 4.446 | 3.505 | 2.742 | 2.028 | 1.460 | 0.997 | 0.633 | 0.356 | 0.165 | 0.051 | -0.016 | -0.059 | -0.072 | -0.081 | -0.084 |
| 8 | 4.441 | 3.517 | 2.821 | 2.121 | 1.481 | 0.961 | 0.579 | 0.303 | 0.123 | 0.017 | -0.039 | -0.066 | -0.081 | -0.085 | -0.085 |
| 9 | 4.428 | 3.547 | 2.829 | 2.162 | 1.550 | 1.014 | 0.592 | 0.282 | 0.088 | -0.015 | -0.068 | -0.090 | -0.098 | -0.097 | -0.094 |
| 10 | 4.395 | 3.481 | 2.748 | 2.043 | 1.453 | 0.981 | 0.615 | 0.340 | 0.151 | 0.041 | -0.026 | -0.060 | -0.081 | -0.087 | -0.089 |
| 11 | 4.390 | 3.513 | 2.778 | 2.095 | 1.478 | 0.963 | 0.569 | 0.283 | 0.100 | -0.006 | -0.059 | -0.085 | -0.097 | -0.095 | -0.095 |
| 12 | 4.409 | 3.480 | 2.683 | 1.938 | 1.360 | 0.914 | 0.559 | 0.301 | 0.127 | 0.025 | -0.035 | -0.064 | -0.082 | -0.088 | -0.089 |
| 13 | 4.427 | 3.499 | 2.724 | 1.988 | 1.377 | 0.888 | 0.522 | 0.254 | 0.094 | 0.001 | -0.053 | -0.079 | -0.087 | -0.090 | -0.090 |
| 14 | 4.388 | 3.490 | 2.792 | 2.071 | 1.397 | 0.871 | 0.498 | 0.238 | 0.075 | -0.015 | -0.064 | -0.087 | -0.097 | -0.095 | -0.093 |
| 15 | 4.419 | 3.533 | 2.771 | 2.067 | 1.424 | 0.885 | 0.489 | 0.218 | 0.053 | -0.033 | -0.074 | -0.092 | -0.097 | -0.097 | -0.095 |
| 16 | 4.421 | 3.474 | 2.671 | 1.945 | 1.355 | 0.889 | 0.537 | 0.276 | 0.109 | 0.013 | -0.043 | -0.071 | -0.087 | -0.090 | -0.089 |
| 17 | 4.292 | 3.374 | 2.616 | 1.916 | 1.354 | 0.899 | 0.556 | 0.299 | 0.124 | 0.021 | -0.038 | -0.070 | -0.086 | -0.090 | -0.092 |
| 18 | 4.489 | 3.506 | 2.722 | 1.947 | 1.254 | 0.730 | 0.375 | 0.146 | 0.018 | -0.051 | -0.083 | -0.095 | -0.098 | -0.096 | -0.093 |
| 19 | 4.394 | 3.520 | 2.713 | 1.921 | 1.242 | 0.740 | 0.394 | 0.164 | 0.024 | -0.042 | -0.079 | -0.089 | -0.096 | -0.095 | -0.092 |
| 20 | 4.364 | 3.481 | 2.716 | 1.932 | 1.219 | 0.702 | 0.366 | 0.150 | 0.018 | -0.048 | -0.081 | -0.096 | -0.099 | -0.098 | -0.095 |
| 21 | 4.401 | 3.503 | 2.725 | 1.879 | 1.101 | 0.551 | 0.233 | 0.037 | -0.053 | -0.092 | -0.109 | -0.113 | -0.109 | -0.103 | -0.099 |

## B. Entropy Rate and Redundancy

To estimate the entropy rate, a polynomial interpolation of third degree is first applied to the values of $F_N$. As an example, Fig. 3 shows the interpolated $F_N$ curves for samples one and twenty-one.

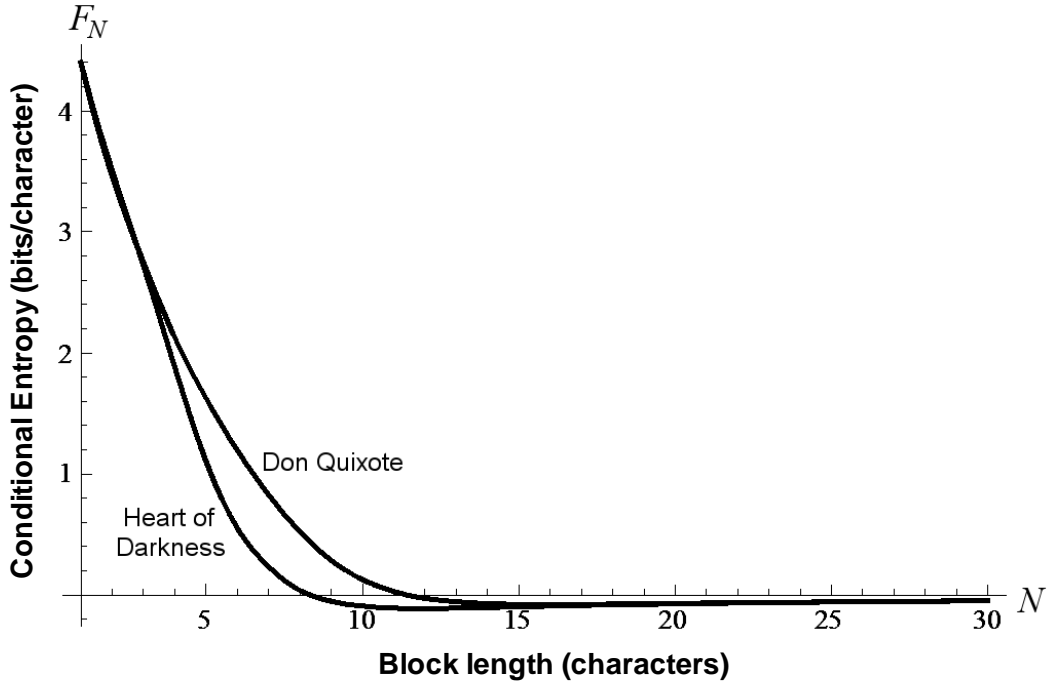

Fig. 3. Interpolated curves of conditional entropy (bits/character) for samples 1 and 21.

As shown in Fig. 3, $F_N$ becomes negative after crossing zero, and then asymptotically approaches zero as $N{\rightarrow}\infty$. Therefore

$$\operatorname*{Lim}_{n \to \infty} F_N = \operatorname*{Lim}_{n \to N_Z} F_N \qquad (5)$$

In (5), $N_Z$ is the root of the continuous function $F_N$ and hence it is a real number. The values of $H_n$ in Table I are similarly mathematically interpolated to find $H_{NZ}$, the real value of $H_n$ corresponding to $N_Z$. The redundancy is then obtained using the classic formula, $R = 1 - \frac{H_L}{H_{max}}$, where $H_L$ is the source's entropy rate, and $H_{max} = \log_2 (A_S)$. The value of $H_L$ is given approximately by

$$H_L \approx \frac{H_{NZ}}{N_Z} \qquad (6)$$

Table V summarizes the values of $N_Z$, $H_{NZ}$, $H_L$, and $R$.

TABLE V
ENTROPY RATE AND REDUNDANCY

| Sample | $N_z$ | $H_{NZ}$ | $H_L$ | $R(\%)$ |
|--------|-------|----------|-------|---------|
| 1 | 11.42 | 17.32 | 1.52 | 75.34 |
| 2 | 11.43 | 17.15 | 1.50 | 76.72 |
| 3 | 11.14 | 17.19 | 1.54 | 75.30 |
| 4 | 11.14 | 17.08 | 1.53 | 75.67 |
| 5 | 9.59 | 17.08 | 1.78 | 72.96 |
| 6 | 10.10 | 16.69 | 1.65 | 73.70 |
| 7 | 10.71 | 16.38 | 1.53 | 75.74 |
| 8 | 10.24 | 16.36 | 1.60 | 74.50 |
| 9 | 9.81 | 16.49 | 1.68 | 73.33 |
| 10 | 10.54 | 16.24 | 1.54 | 75.11 |
| 11 | 9.92 | 16.17 | 1.63 | 73.75 |
| 12 | 10.35 | 15.79 | 1.53 | 75.43 |
| 13 | 10.01 | 15.77 | 1.58 | 74.28 |
| 14 | 9.78 | 15.81 | 1.62 | 74.04 |
| 15 | 9.52 | 15.85 | 1.66 | 73.02 |
| 16 | 10.19 | 15.69 | 1.54 | 74.79 |
| 17 | 10.29 | 15.44 | 1.50 | 76.11 |
| 18 | 9.21 | 15.18 | 1.65 | 73.91 |
| 19 | 9.28 | 15.11 | 1.63 | 73.96 |
| 20 | 9.21 | 14.94 | 1.62 | 73.96 |
| 21 | 8.31 | 14.42 | 1.73 | 71.60 |

In Table V, the weighted average of $H_L$ is 1.58 bits/character. The weighted average of the alphabet size in Table I is 77.91 characters. Therefore the average redundancy, $R$, for the set of twenty-one samples, comprising nearly 20.3 million characters, is approximately:

$$R = 1 - \frac{1.58}{\{\log_2 77.91\}} \approx 74.86\%$$

Naturally, this value of $R$ would be precise only for the set of works analyzed in this study. However, other samples of written English should have a redundancy value of about 74.86%, as long as their length is reasonably long. This method for finding the entropy can be done in very reasonable time since the roots of $F_N$ occur at small values of $n$. Textbooks on Information Theory generally consider 75% to be the redundancy of English [9], but arrive at such a value through different assumptions and values for both $H_L$ and $A_S$.

In Table V, it can also be observed that the sample with the lowest WDR has the highest redundancy, and vice versa. Samples five and seventeen are another example of the same behavior.

*C. Word equiprobability distance*

The equiprobability distance, $d_{aep}$, in words, can be approximated by

$$d_{aep} \approx \frac{n_{aep}}{\alpha+1} \qquad (7)$$

Table VI shows the values of $d_{aep}$ on every sample from $n$ = 1 to 500 characters; $d_{aep}$ provides information about the number of words after which the structural correlation between words disappears. Table VI also shows that $2^{n_{aep}H_L}$, the number of typical sequences of length $n_{aep}$ characters, is of considerable size despite the apparently small number of words involved.

TABLE VI
EQUIPROBABLE DISTANCE AND
TYPICAL SEQUENCES

| Sample | $d_{aep}$ | $2^{n_{aep}H_L}$ |
|--------|-----------|-------------------|
| 1 | 25.30 | 2.24E+59 |
| 2 | 9.32 | 1.32E+22 |
| 3 | 14.41 | 8.07E+33 |
| 4 | 8.84 | 4.96E+21 |
| 5 | 18.46 | 1.16E+53 |
| 6 | 15.49 | 1.97E+41 |
| 7 | 11.83 | 1.20E+28 |
| 8 | 13.37 | 1.57E+33 |
| 9 | 11.04 | 6.99E+28 |
| 10 | 7.11 | 4.22E+17 |
| 11 | 58.80 | 1.33E+151 |
| 12 | 7.63 | 6.78E+18 |
| 13 | 8.29 | 2.53E+20 |
| 14 | 6.58 | 1.09E+17 |
| 15 | 5.22 | 3.37E+13 |
| 16 | 7.05 | 4.87E+16 |
| 17 | 6.69 | 2.33E+15 |
| 18 | 9.14 | 6.85E+22 |
| 19 | 5.84 | 4.95E+14 |
| 20 | 6.70 | 3.78E+17 |
| 21 | 14.47 | 1.48E+39 |

Language processing tools such as plagiarism detection software should take into account the value of $n_{aep}$, before passing judgment, because for sequences shorter than $n_{aep}$, there is an important statistical influence imposed by the structural relation between words.

## V. Conclusion

There are innumerable contexts in which the English language is, has been, and will continue to be used, perhaps as many different contexts as in anyone's life. However, several statistical properties of the language can be identified when reasonably long samples are analyzed.

A simple method for finding the entropy and redundancy of sample of English language has been presented. For the nearly 20.3 million printable characters of English text analyzed in this work, an entropy rate of 1.58 bits/character was found, and a language redundancy of 74.86%. For the set of samples analyzed, the probability of a typical sequence of length $n$ would then be $2^{-n\,(1.58)}$, assuming $n$ is sufficiently large.

The maximum values of entropy for $n$-character symbols for the set of samples analyzed were found to occur for $n$ between 8.31 and 11.43 characters (approximately 1.54 and 2.17 words). Although the analysis has been done for symbols from $n = 1$ to 500 characters in length, it has been found that only this maximum value of entropy is required for finding the entropy rate of the sample. Also, entropy values for different shifts for the same $n$ were found to be very similar.

Because of the structural relationship between words that is imposed by language, tools such as plagiarism detection software should take into consideration the value of $d_{aep}$ before attempting a verdict. The same applies, for example, to the limit on number of words imposed by some social networks on the Internet for short messages. Intuitively, it sounds obvious that there should be sufficient words in a short text message to allow a meaning to be conveyed. What $d_{aep}$ indicates is that, in general, such a number is fairly low. However, finding the exact value of $d_{aep}$ for a given sample may require a considerable computing time.

It is difficult to talk about a definitive and precise value of the entropy rate of a language because the condition that $n \rightarrow \infty$ requires an extremely large sample to be analyzed, which in practice simply is not available, apart from it taking a very long time to compute. In this work, however, an approach to estimating both the entropy and redundancy of a reasonably large language sample in a simple, direct way has been presented. Computer processing proved not to be an insurmountable barrier in the analysis of this type of sample and, on the contrary, allowed to corroborate interesting observations about the entropy of English.

REFERENCES

[1]  C. E. Shannon, "Prediction and entropy of printed English," *Bell Syst. Tech. J.*, vol. 30, pp. 47–51, Jan. 1951.

[2]  T. M. Cover and R. C. King, "A convergent gambling estimate of the entropy of English," *IEEE Trans. Inform. Theory*, vol. IT-24, pp. 413–421, July 1978.

[3]  W. J. Teahan, and J. G. Cleary, "The entropy of English using PPM-based models," in  *Proc. Data Compression Conference*, IEEE Soc. Press, pp 53–62, 1996.

[4]  I. Kontoyiannis, "The Complexity and Entropy of Literary Styles," *NSF Technical Report No. 97*, Department of Statistics, Stanford University, 1996.

[5]  O. A. Rosso, H. Craig, and P. Moscato, "Shakespeare and other English Renaissance authors as characterized by Information Theory complexity quantifiers," *Physica A: Statistical Mechanics and its Applications*, vol. 388, Issue 6, pp. 916–926, Mar. 2009.

[6]  F. G. Guerrero. Supporting Material for "A New Look to Classic Entropy of Written English". [Online]. Available at: http://sistel-uv.univalle.edu.co/EWE.html.

[7]  The Daily Telegraph, "100 novels everyone should read", January 16 2009. [Online]. Available: http://www.telegraph.co.uk/culture/books/4248401/100-novels-everyone-should-read.html. Accessed on October 29, 2009.

[8]  The Gutenmberg Project. [Online]. Available: http://www.gutemberg.org. Accessed on August 13, 2009.

[9]  R. B. Wells, *Applied Coding and Information Theory for Engineers*. Upper Saddle River, New Jersey: Prentice Hall, ch. 8, 2002.

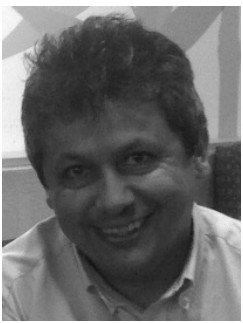

**Fabio G. Guerrero**, received a B. Eng. degree in telecommunications engineering from Universidad del Cauca, Popayán, Colombia (South America), in 1992, and a M.Sc. degree in Real-Time Electronic Systems from Bradford University, UK, in 1995. Currently, he works as telecommunications assistant lecturer in the Department of Electrical and Electronics Engineering of Universidad del Valle, Cali, Colombia (South America). His academic interests include digital communications, telecommunication systems modeling, and next generation networks. He is a member of the IEEE Communications Society.